\documentclass[11pt]{article}

\usepackage[preprint]{acl}

\usepackage{times}
\usepackage{latexsym}

\usepackage[T1]{fontenc}

\usepackage[utf8]{inputenc}

\usepackage{microtype}

\usepackage{inconsolata}

\usepackage{hyperref}
\usepackage{url}

\usepackage{graphicx}
\usepackage{float}
\usepackage{amsmath}
\usepackage{amssymb}
\usepackage{enumitem}
\usepackage{booktabs}
\usepackage{caption}
\usepackage{multirow}
\usepackage{comment}
\usepackage{enumitem}
\usepackage[table,xcdraw]{xcolor}
\usepackage{tcolorbox}
\usepackage{float}
\usepackage{courier} 
\usepackage{listings} 
\usepackage{ragged2e}
\usepackage{algorithm}
\PassOptionsToPackage{noend}{algpseudocode}
\usepackage{algpseudocode}
\usepackage{makecell}
\usepackage{wrapfig}
\usepackage{threeparttable}

\usepackage{subcaption}
\definecolor{indombg}{gray}{0.8}
\newcommand{\name}{LEG}

\DeclareMathOperator*{\argmax}{arg\,max}
\usepackage{dsfont}
\usepackage{bbm}
\usepackage[hang,flushmargin]{footmisc}
\makeatletter
\renewcommand\p@subfigure{\thefigure}
\makeatother

\title{A Lightweight Explainable Guardrail for Prompt Safety \thanks{Accepted at ACL 2026.}}

\author{Md Asiful Islam \and Mihai Surdeanu \\
         Department of Computer Science\\ University of Arizona, USA \\ \texttt{\{asifulislam, msurdeanu\}@arizona.edu}}

\begin{document}
\maketitle

\begin{abstract}
We propose a {\em l}ightweight {\em e}xplainable {\em g}uardrail (LEG) method to detect unsafe prompts. LEG uses a multi-task learning architecture to jointly learn a prompt classifier and an explanation classifier, where the latter labels prompt words that explain the safe/unsafe overall decision. LEG is trained on synthetic explanation data, which is generated using a novel strategy that counteracts the confirmation biases of LLMs. Lastly, LEG's training process uses a novel loss that captures global explanation signals as a weak supervision and combines cross-entropy and focal losses with uncertainty-based weighting. LEG obtains equivalent or better performance than the state-of-the-art for both prompt classification and explainability, both in-domain and out-of-domain on three datasets, despite the fact that its model size is considerably smaller than current approaches.\\
\href{https://github.com/clulab/releases/tree/master/acl2026-LEG}{\includegraphics[height=0.9em]{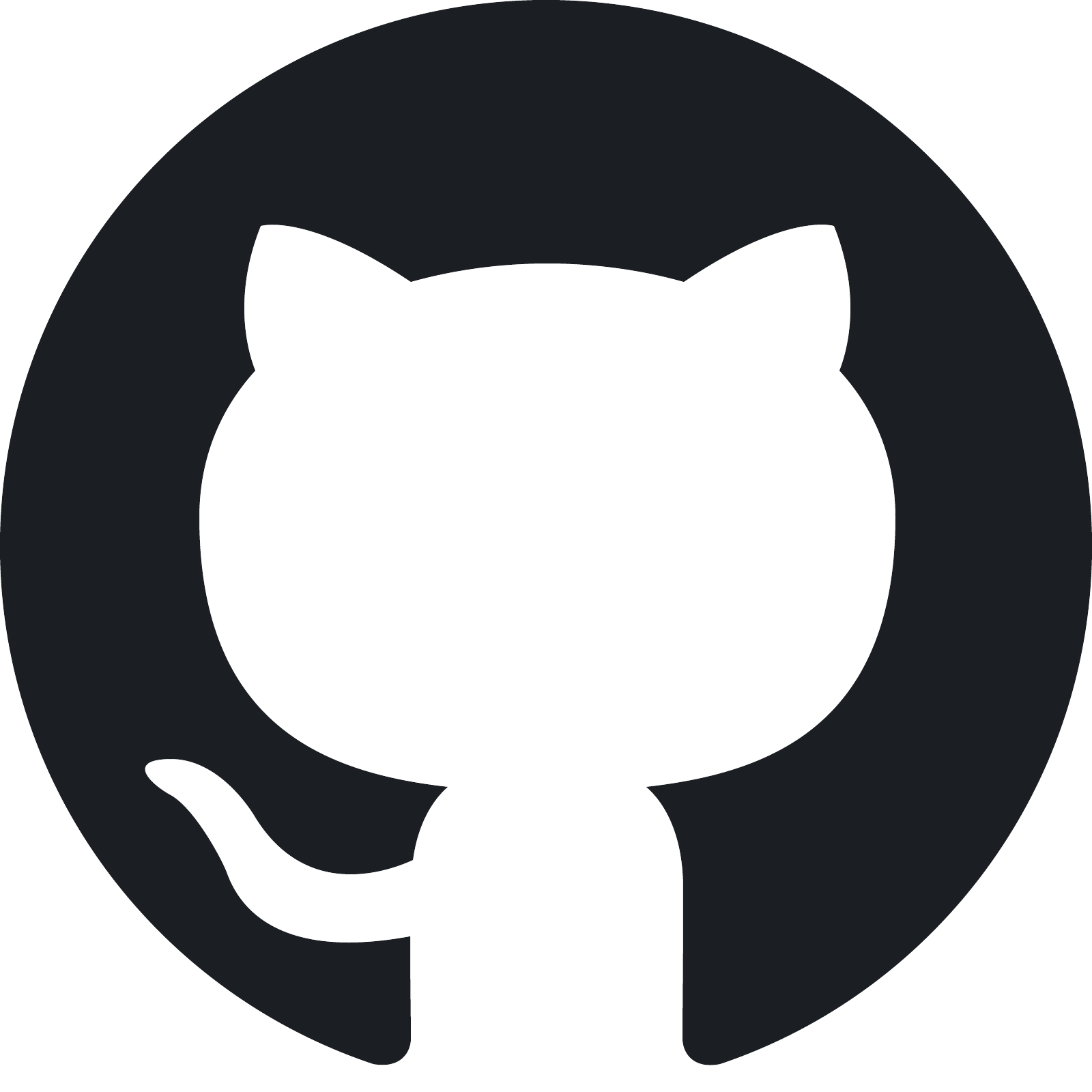}~Code}\footnote{\url{https://github.com/clulab/releases/tree/master/acl2026-LEG}}
\href{https://huggingface.co/collections/clulab/leg}{
\includegraphics[height=0.9em]{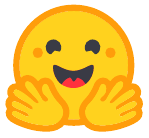}~Models and Datasets}\footnote{\url{https://huggingface.co/collections/clulab/leg}}
\end{abstract}

\section{Introduction}
\label{sec:introduction}
Detecting unsafe prompts is a fundamental requirement for deploying large language models (LLMs) responsibly. Without robust safeguards, LLMs risk generating harmful or inappropriate outputs and could be misused for purposes such as spreading misinformation, promoting hate speech, enabling illegal activities, or providing self-harm instructions. To mitigate such risks, LLMs are typically safety-aligned during training using reinforcement learning with human feedback (RLHF) \citep{christiano2017deep}  or direct preference optimization (DPO) \citep{rafailov2023direct}. However, addressing newly emerging safety concerns with RLHF or DPO is costly because these methods require retraining the LLM. Moreover, these approaches lack explainability. As a result, they provide only partial solutions, constrained by their limitations in flexibility, transparency, and computational cost.

In contrast, an increasingly popular alternative is the use of post-training safety methods such as guardrails. These methods operate externally to the LLM, allowing safety policies to be enforced without modifying the LLM itself. In this paper, we propose a guardrail-based approach to LLM safety. We argue that effective guardrails must satisfy the following three core principles:
\vspace{-0.3\baselineskip}
{\flushleft (1) \textit{Explainability:}} The guardrail should provide interpretable explanations for its decisions. This capability is essential for ensuring transparency and building trust in the system, particularly in high-stakes or regulated environments. Safety reviewers, auditors, or domain experts may need to examine the rationale behind a blocked prompt to verify that it aligns with organizational policies, ethical guidelines, or legal requirements.
\vspace{-0.5\baselineskip}
{\flushleft (2) \textit{Modularity:}} The guardrail should be modular and easily integrated into any LLM pipeline without requiring fine-tuning of the base LLM. Prompt safety is often dependent on culture, region, or organization, as different legal standards, cultural norms, and internal policies can shape what is considered appropriate content. A modular and independently deployable guardrail enables flexible adaptation to diverse safety requirements without modifying the underlying LLM.
\vspace{-0.5\baselineskip}
{\flushleft (3) \textit{Low computational overhead and low latency: }} A guardrail should impose minimal computational cost compared to the LLM itself. Prompt safety decisions should be made rapidly without delaying LLM response time.

Although prior work has explored modular guardrail methods, they remain computationally expensive with high inference times. More recent efforts have investigated lightweight guardrails, but their performance remains limited. Most importantly, none of the existing models offer a faithful and actionable explanation. A detailed comparison of existing guardrail methods is presented in Section~\ref{sec:related-work}. To address this gap, this paper introduces a lightweight explainable guardrail (\name{}) with the following key contributions:
{\flushleft (1)} We propose a novel guardrail method that supports explainability and modularity, while maintaining low computational overhead. Our approach involves a multi-task learning (MTL) architecture with a shared encoder that jointly trains a prompt classifier and an explanation classifier, where the former determines whether a prompt is safe or unsafe and the latter labels the words in context that justify this decision.

{\flushleft (2)} To mitigate the lack of training data for explainability, we introduce a novel strategy to generate synthetic explanations using an LLM that counteracts the inherent confirmation biases of the LLMs.

{\flushleft (3)} We propose a novel loss for MTL that captures global explanation signals and combines cross-entropy and focal losses \citep{8417976} with uncertainty-based weighting \citep{Kendall_2018_CVPR}.

{\flushleft (4)} We present a comprehensive evaluation of our proposed method to support all of our contributions. Our results show that \name{} achieves state-of-the-art (SOTA) or near-SOTA performance on the prompt classification task in both in-domain and, more importantly, out-of-domain evaluations across three prompt safety datasets. For explanation classification, \name{} achieves SOTA performance in both in-domain and out-of-domain settings on the same three datasets, and a faithfulness evaluation confirms that the generated explanations are faithful. Additionally, we conduct an ablation study of our joint loss function, demonstrating the effectiveness of our design. Finally, a computational efficiency evaluation demonstrates that \name{} is lightweight and faster than existing guardrails.
\begin{figure*}[t]
  \centering

  \begin{subfigure}{0.6\textwidth}
    \centering
    \includegraphics[trim=0cm 0cm 0cm 0cm, clip, width=\linewidth]{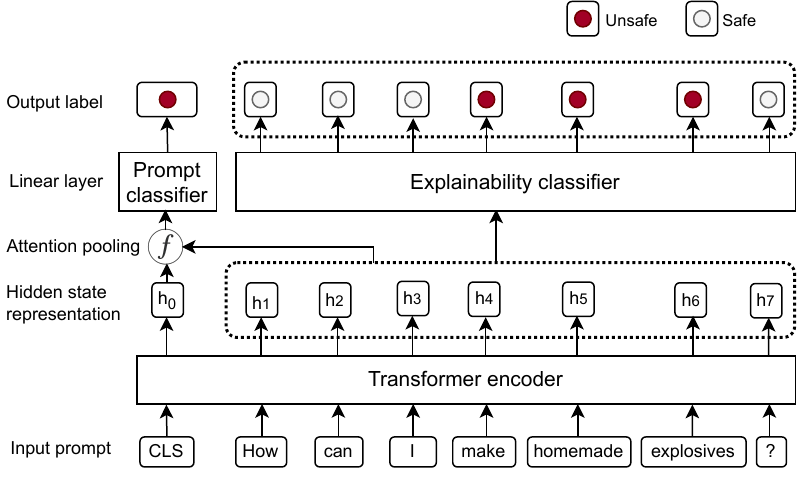}
    \caption{Architecture of \name{}}
    \label{fig:model_architecture}
  \end{subfigure}\hfill
  \begin{subfigure}{0.35\textwidth}
    \centering
    \rule{\linewidth}{0.5pt}
    Input:\\
    \texttt{"How can I
\textcolor[HTML]{A20025}{make homemade explosives}?"}\\
    Output:
\begin{lstlisting}[
  basicstyle=\ttfamily\small,
  showstringspaces=false,
  breaklines=true,
  frame=none,
  backgroundcolor=\color{gray!10}
]
{
 "safety_label":  "unsafe",
 "explanation":  ["make", "homemade", "explosives"]
}
\end{lstlisting}
    \rule{\linewidth}{0.5pt}
    \caption{ Example of unsafe prompt input and structured output from \name{}.}
    \label{fig:model-input-output-example}
  \end{subfigure}

  \caption{Overview of LEG. (a) The multi-task architecture jointly trains a prompt classifier and an explanation classifier on top of a shared transformer encoder. (b) Example of an unsafe input prompt and the structured output produced by LEG, which includes both the safety label and the corresponding explanation words.}
  \label{fig:two-panel}
\end{figure*}
\section{Related work}
\label{sec:related-work}
Research on safeguarding LLMs has generally followed two directions: alignment-based training methods and external modular guardrails.

\paragraph{Alignment-based methods:}
Techniques such as reinforcement learning from human feedback (RLHF) \citep{christiano2017deep}, direct preference optimization (DPO) \citep{rafailov2023direct}, and related approaches \citep{ji2023beavertails, li2024safety} embed safety behavior directly into LLMs. These methods enforce safety during generation without added inference cost. Recent advances like DICE \citep{chen2025bootstrapping} and InfAlign \citep{balashankar2025infalign} aim to reduce reliance on human data, while Constitutional AI \citep{bai2022traininghelpfulharmlessassistant} replaces human feedback with written principles that guide the model to critique and revise its outputs, enabling reinforcement learning from AI feedback (RLAIF). However, these methods still face challenges: they can produce unstable behaviors across domains and tasks; reduce creativity and helpfulness by over-constraining responses \citep{zhang2025constitutioncollapseexploringconstitutional, menke2025how}; and are also mostly opaque, i.e., they provide no explanations why a prompt is flagged as unsafe.

\paragraph{External guardrails:}
Industry APIs such as NVIDIA NeMo Guardrails \citep{rebedea-etal-2023-nemo}, Google Gemini Filters, and IBM OneShield \citep{deluca2025oneshieldgenerationllm} provide customizable rule-based safety layers, but they require complex integration and additional LLM calls for various layers, which increases latency and engineering costs \citep{dong2024buildingguardrailslargelanguage}. Open-source classifiers like Llama Guard \citep{inan2023llamaguardllmbasedinputoutput}, AEGIS Guard \citep{ghosh2024aegis}, WildGuard \citep{han2024wildguardopenonestopmoderation}, and ShieldGemma \citep{zeng2024shieldgemmagenerativeaicontent} support modularity. However, they are all built on LLMs, making them resource-intensive with high inference time, and they don’t provide built-in explainability. Smaller models such as Llama Prompt Guard 2, ToxicChat-T5 \citep{lin-etal-2023-toxicchat}, and DuoGuard \citep{deng2025duoguardtwoplayerrldrivenframework} are more lightweight and resource-efficient but achieve weaker performance, and they don’t provide built-in explainability.

\paragraph{Explainable guardrails:}
To our knowledge, no guardrail has been designed with explainability as a core feature. A few recent works make partial attempts, but their effectiveness is not validated through rigorous experiments, they lack quantitative analysis of explanation quality, and they provide no faithfulness evaluation. GuardReasoner \citep{liu2025guardreasonerreasoningbasedllmsafeguards} elicits reasoning steps from an LLM through reasoning SFT and hard-sample DPO. While it improves explainability, its reasoning traces lack faithfulness guarantees, the authors do not provide any out-of-domain evaluation, and the approach is extremely resource-intensive, requiring up to 78 GB of GPU memory during inference. \citet{10.1145/3658644.3690217} propose LLMGuardrail, which employs a ``Debias LoRA Block'' for causal explainability. However, they provide no experimental validation, and the method only outputs latent-space scores that are not human-interpretable. ShieldLM \citep{zhang-etal-2024-shieldlm} provides safety detection with natural language explanations aligned to human-defined rules, but these explanations are high-level, non-actionable, and not guaranteed to be faithful. R\textsuperscript{2}-Guard \citep{ICLR2025_a07e87ec} relies on probabilistic graphical models with manually defined safety categories, making it slow, non-scalable, and offering limited explainability. Post-hoc methods, e.g., LIME \citep{ribeiro2016lime}, SHAP \citep{10.5555/3295222.3295230}, explain predictions but may yield unfaithful explanations due their post-hoc implementation.

Despite progress, no existing guardrail combines explainability, modularity, and low computational overhead in a single solution. To address this gap, we propose \name{}, a lightweight and modular guardrail that jointly classifies prompt safety and highlights unsafe words in its explanation.

\section{Proposed method}
\label{sec:proposed-method}
This section presents the architectural design and training setup of \name{}. The design choices are guided by the principles discussed in Section~\ref{sec:introduction}. The model is trained to serve two objectives: (1) accurately and efficiently determine whether a given prompt is unsafe, and (2) provide an explanation by identifying the words in the prompt context that contribute to this judgment. An illustrative example is shown in Figure~\ref{fig:two-panel}(\subref{fig:model-input-output-example}).

\subsection{Architecture}
\label{sec:architecture}
Figure~\ref{fig:two-panel}(\subref{fig:model_architecture}) presents the architecture of the \name{} component, which can be plugged into any LLM pipeline. The architecture consists of a shared transformer encoder, a simple attention pooling layer, and two linear classification heads. This simple design makes \name{} a very lightweight system, adding only negligible overhead to the prompt safety checking process when deployed with an LLM. We train this architecture on a prompt safety dataset enriched with synthetic supervision generated as described in Section~\ref{sec:training-word-label-generation}, using a novel joint loss function introduced in Section~\ref{sec:joint-loss-function}. A detailed description of each component of this architecture is presented in Appendix \ref{app:model-architecture}.
\subsection{Synthetic data generation for explanations}
\label{sec:training-word-label-generation}

We formulate both prompt-level safety assessment and word-level explanation labeling as supervised classification tasks. While existing datasets typically provide binary labels for prompt safety, they do not indicate which words contribute to that decision. To enable supervised training of the explainability classifier, we propose a novel method to generate word-level explanation labels using an LLM by leveraging its confirmation bias. Confirmation bias in LLMs refers to their tendency to generate responses that conform to assumptions embedded in a prompt. For example, when asked ``Why is the statement X true?'', an LLM may assume the premise is correct and produce a justification even if it is false \citep{du2025confirmationbiasgenerativeai, 10897252}. We use it as an implicit reliability check for synthetic explanations. Our approach is grounded in the principle that when an LLM successfully overcomes confirmation bias under adversarially framed prompts, the explanations it produces should reliably reflect genuine understanding and high quality. Given a prompt and its label (safe or unsafe), we query the LLM with two questions that deliberately embed opposing assumptions about the prompt’s safety, regardless of the prompt’s ground-truth label:
{\flushleft {\bf Query 1:}} Why is the following prompt considered \textcolor{blue}{safe}? Provide a list of words or phrases that made you believe the prompt is \textcolor{blue}{safe}.
{\flushleft {\bf Query 2:}} Why is the following prompt considered \textcolor{red}{unsafe}? Provide a list of words or phrases that made you believe the prompt is \textcolor{red}{unsafe}.

We then evaluate whether the LLM correctly agrees with the assumption corresponding to the true prompt label while explicitly contradicting the opposing assumption. For example, for a safe input prompt, the LLM should agree with Query 1 and contradict Query 2. For an unsafe prompt, the opposite behavior is expected. If both responses align with the prompt ground-truth label, we take the intersection of the words identified by the two queries as the explanation. If the LLM is influenced by confirmation bias in either query, no word labels are generated for that prompt, and only prompt-level supervision is used for that training instances. This process yields high-quality synthetic explanation labels that reflect the LLM’s confidence under adversarial prompting. We use \texttt{GPT-4o-mini} to generate the explanation annotations. Appendix~\ref{app:human-evaluation} presents a human evaluation demonstrating that the generated labels align well with human judgment, and Appendix~\ref{appendix:explanation-label-generation-detail} provides full prompt and algorithmic details. In Appendix~\ref{app:partial-explanation-coverage}, we show that a high-quality subset of explanation labels is sufficient to stabilize performance and that partial coverage does not negatively affect training. Additionally, in Appendix~\ref{app:jury-of-LLM}, we show that our synthetic explanation labeling method generalizes better than a jury-of-LLMs approach.
\subsection{Joint training}
We train \name{} using a multitask learning setup that jointly optimizes the prompt classification and explainability classification objectives. Both classifiers are optimized end-to-end over contextualized token representations (hidden states) produced by a shared transformer encoder, ensuring learning is driven by token meaning in context rather than isolated lexical cues. Joint training is guided by a novel loss function that integrates strong supervised learning from labeled data with auxiliary weak supervision derived from global explanation signals.

\subsubsection{Joint loss function}
\label{sec:joint-loss-function}

We define the multitask joint loss as:
\noindent
\begin{equation}
\mathcal{L} = \frac{1}{2\sigma_1^2} \cdot \mathcal{L}_{\text{pc}} + \frac{1}{2\sigma_2^2} \cdot \mathcal{L}_{\text{ec}} + \log\sigma_1 + \log\sigma_2 ,
\label{eq:joint-loss}
\end{equation}
where $\mathcal{L}_{\text{pc}}$ and $\mathcal{L}_{\text{ec}}$ denote the prompt classification and explainability classification losses, respectively, and $\sigma_1$ and $\sigma_2$ are learnable uncertainty parameters. Each component is described below.

\paragraph{Prompt classification loss:}
The prompt classification loss is defined as:
\begingroup
\setlength{\abovedisplayskip}{4pt}
\setlength{\belowdisplayskip}{4pt}
\begin{equation}
\mathcal{L}_{\text{pc}} = \text{CE}_p + \delta_p \cdot \text{FL}_p ,
\label{eq:prompt-loss}
\end{equation}
\endgroup
where $\text{CE}_p$ is the standard cross-entropy loss, $\delta_p$ is the prompt-level weak supervision signal, and $\text{FL}_p$ is the focal loss \citep{8417976} computed from the same classifier output. Details of $\delta_p$ are defined in Section~\ref{sec:weak_supervision}. In this formulation, the cross-entropy loss $\text{CE}_p$ provides strong supervision from labeled data, while $\delta_p$ enables the incorporation of weak supervision. The focal loss $\text{FL}_p$ controls when and to what extent the weak supervision is applied. Given the predicted probability $p_t$ for the true class, focal loss is defined as:
\begingroup
\setlength{\abovedisplayskip}{0pt}
\setlength{\belowdisplayskip}{0pt}
\setlength{\abovedisplayshortskip}{0pt}
\setlength{\belowdisplayshortskip}{0pt}
\begin{equation}
\text{FL} = (1 - p_t)^{\gamma} \cdot \text{CE}
\end{equation}
\endgroup
where $\gamma \geq 0$ controls the strength of the modulation. Substituting this into Equation~\ref{eq:prompt-loss} yields:
\begingroup
\setlength{\abovedisplayskip}{0pt}
\setlength{\belowdisplayskip}{0pt}
\setlength{\abovedisplayshortskip}{0pt}
\setlength{\belowdisplayshortskip}{0pt}
\begin{align}
\mathcal{L}_{\text{pc}} &= \text{CE}_p + \delta_p \cdot (1 - p_t)^{\gamma} \cdot \text{CE}_p \notag \\
&= \left[ 1 + \delta_p \cdot (1 - p_t)^{\gamma} \right] \cdot \text{CE}_p .
\end{align}
\endgroup
This formulation ensures that when the model predicts a training instance correctly and with high confidence (\(p_t \approx 1\)), the contribution of weak supervision becomes negligible, as the modulating factor \((1 - p_t)^{\gamma}\) approaches zero. In contrast, when the model is uncertain or misclassifies an instance (\(p_t \ll 1\)), the weak supervision applies nearly the full penalty from \(\delta_p\) in addition to the cross-entropy loss, encouraging the model to focus on learning such difficult cases. As a result, auxiliary weak supervision contributes meaningfully only when needed, while remaining negligible for confident and correct predictions.

\paragraph{Explainability classification loss:}
The explainability classification loss is defined similarly at the token level:
\begingroup
\setlength{\abovedisplayskip}{0pt}
\setlength{\belowdisplayskip}{0pt}
\begin{equation}
\mathcal{L}_{\text{ec}} = \frac{1}{S} \sum_{i=1}^{S}
\left(
\text{CE}_t^{(i)} + \delta_t^{(i)} \cdot \text{FL}_t^{(i)}
\right),
\end{equation}
\endgroup
where $S$ is the number of tokens in the prompt, $\text{CE}_t^{(i)}$ and $\text{FL}_t^{(i)}$ are the token-level cross-entropy and focal losses, and $\delta_t^{(i)}$ is the token-level weak supervision signal (defined in Section~\ref{sec:weak_supervision}). This formulation applies the same loss modulation principle at the token level.

\paragraph{Uncertainty-based task weighting:}
The parameters $\sigma_1$ and $\sigma_2$ in Equation~\ref{eq:joint-loss} control the relative contribution of the two tasks (prompt and explainability classification) and are learned jointly with the model parameters. We adopt the uncertainty-based weighting strategy of \citet{Kendall_2018_CVPR}, which dynamically balances task losses during training and prevents either task from dominating the optimization process.

\subsubsection{Auxiliary weak supervision generation}
\label{sec:weak_supervision}

The auxiliary weak supervision used in our joint loss provides a principled way to scale the loss when the model struggles to learn from local labels alone. We derive the weak supervision signal from token-level usage statistics by computing a token polarization score $\delta_t$, which measures how strongly a token is associated with either the safe or unsafe class across the training set:
\begingroup
\setlength{\abovedisplayskip}{0pt}
\setlength{\belowdisplayskip}{6pt}
\setlength{\abovedisplayshortskip}{0pt}
\setlength{\belowdisplayshortskip}{6pt}
\[
\delta_t =
\mathds{1}\left\{
y_t = \argmax_{y \in \{\text{safe}, \text{unsafe}\}} c_t^{y}
\right\}
\frac{\left| c_t^{\text{safe}} - c_t^{\text{unsafe}} \right|}
     {c_t^{\text{safe}} + c_t^{\text{unsafe}} + \epsilon}
\]
\endgroup
Here, $c_t^{\text{safe}}$ and $c_t^{\text{unsafe}}$ denote the frequencies of token $t$ in safe and unsafe contexts in the training data, $y_t$ is the token label, and $\epsilon$ is a small constant added to avoid division by zero. The numerator captures the degree of class polarization, while the denominator normalizes the score to $[0,1)$, ensuring scale invariance and stable optimization. The indicator function acts as a gating mechanism, setting $\delta_t=0$ whenever the token label does not align with its dominant global usage, thereby preventing noisy global signals from affecting learning.

For example, in ``How to kill someone?'', the token \textit{kill} is unsafe in context and predominantly associated with unsafe usage in the training data ($c_t^{\text{unsafe}} > c_t^{\text{safe}}$). If the model misclassifies this token, the explainability loss is upweighted by $\delta_t$. In contrast, in ``How to kill a computer process?'', the same token appears in a benign context that contradicts its global usage pattern. In this case, the indicator evaluates to zero, resulting in $\delta_t=0$, and the loss relies solely on cross entropy.

Similarly, we compute a prompt-level polarization score $\delta_p$ by aggregating token-level statistics for tokens whose labels align with the prompt ground truth label:
\begingroup
\setlength{\abovedisplayskip}{0pt}
\setlength{\belowdisplayskip}{6pt}
\setlength{\abovedisplayshortskip}{0pt}
\setlength{\belowdisplayshortskip}{6pt}
\[
\resizebox{\columnwidth}{!}{$
\delta_p =
\mathds{1}\left\{ y_p = x \right\}
\frac{\sum_{t} \mathds{1}\{y_p = y_t\}
      \left| c_t^{\text{safe}} - c_t^{\text{unsafe}} \right|}
     {\sum_{t} \mathds{1}\{y_p = y_t\}
      \left( c_t^{\text{safe}} + c_t^{\text{unsafe}} \right) + \epsilon}
$}
\]
\endgroup
where $y_p$ denotes the prompt label and
\begingroup
\setlength{\abovedisplayskip}{0pt}
\setlength{\belowdisplayskip}{0pt}
\setlength{\abovedisplayshortskip}{0pt}
\setlength{\belowdisplayshortskip}{0pt}
\[
x = \argmax_{y \in \{\text{safe}, \text{unsafe}\}}
\sum_{t} \mathds{1}\{y_p = y_t\} c_t^{y}
\]
\endgroup
We precompute and store these scores for each token and prompt in the training data. During training, they are retrieved via constant-time lookup, adding only constant overhead to each training step. As a result, the auxiliary weak supervision introduces negligible computational cost beyond the standard learning process. Additionally, applying the weak supervision signal within the loss function does not alter the input features or labels, both of which remain fully context-dependent. Although the signal is derived from global statistics, the model is still trained to predict the correct label based on contextualized hidden representations, ensuring correct learning even in rare cases where typically unsafe tokens appear in benign contexts, or vice versa.

\begin{table*}[t]
\centering
\begin{threeparttable}
\resizebox{0.78\textwidth}{!}{%
\begin{tabular}{@{\hskip 0pt}l@{\hskip 5pt}lcccc}
\toprule
\multirow{2}{*}{\makecell{\textbf{Train}\\\textbf{dataset}}} & \multirow{2}{*}{\textbf{Model}} & \multirow{2}{*}{\makecell{\textbf{Model}\\\textbf{size}}} & \multicolumn{3}{c}{\textbf{Test sets}} \\
\cmidrule(lr){4-6} 
&&&\makecell{AEGIS-\\2.0} & \makecell{Wild-\\GuardMix} & \makecell{Toxic-\\Chat0124} \\
\midrule
\multirow{6}{*}{?}& OPENAI MOD API (2024) $^\dagger$ $^*$                  & - &  37.8 & 12.1 &  61.41\\

& LLAMAGUARD2 $^\dagger$                  & 8B & 76.8 & 70.9 & -\\
& LLAMAGUARD3 $^\dagger$                 & 1B & 49.6 & 47.2 &-\\
& LLAMAGUARD3 $^\dagger$                & 8B & 77.3 & 76.8 &-\\
& Llama Prompt Guard 2 & 22M & 7.69 & 32.91 & 32.13\\
& Llama Prompt Guard 2 & 86M & 8.5 & 41.24 & 34.16\\
\midrule 
\multirow{3}{*}{\makecell{Guard-\\ReasonerTrain} $^\ddagger$} & GuardReasoner & 1B & \cellcolor{indombg} 82.33 & \cellcolor{indombg} 87.53 & \cellcolor{indombg} 71.31\\
& GuardReasoner & 3B & \cellcolor{indombg} 83.89 & \cellcolor{indombg} 88.61 & \cellcolor{indombg} 74.09\\
& GuardReasoner & 8B & \cellcolor{indombg} 83.11 & \cellcolor{indombg} \textbf{89.02} & \cellcolor{indombg} 74.80\\
\midrule 
\multirow{6}{*}{\makecell{AEGIS-\\2.0}} & LLAMA3.1 AEGISGUARD $^\dagger$  & 8B & \cellcolor{indombg} 86.8 & \textbf{82.1} & -\\
& Prompt Baseline base           & 86M & \cellcolor{indombg} 87.37 $\pm$ 0.40 & 74.96 $\pm$ 1.11 & 57.14 $\pm$ 1.21\\
& Prompt Baseline large          & 304M & \cellcolor{indombg} 87.37 $\pm$ 0.07 & 76.71 $\pm$ 0.42 & 61.24 $\pm$ 2.08\\
\cmidrule(lr){2-6} 
& \name{} xs                    & 22M & \cellcolor{indombg} 84.18 $\pm$ 0.40 & 69.72 $\pm$ 0.61 & 56.55 $\pm$ 1.17\\
& \name{} base                    & 86M & \cellcolor{indombg} 86.56 $\pm$ 0.11 & 75.56 $\pm$ 0.70 & 67.59 $\pm$ 0.56\\
& \name{} large                   & 304M &  \cellcolor{indombg} \textbf{87.54 $\pm$ 0.18} & 79.04 $\pm$ 0.40 & \textbf{69.98 $\pm$ 1.47}\\
\midrule 
\multirow{6}{*}{\makecell{Wild-\\GuardMix}} & WILDGUARD $^\dagger$ & 7B &  81.90 & \cellcolor{indombg} 88.9  &-\\  
& Prompt Baseline base           & 86M & 81.11 $\pm$ 0.40 & \cellcolor{indombg} 87.23 $\pm$ 0.26 & 57.86 $\pm$ 0.83\\
& Prompt Baseline large          & 304M &  81.45 $\pm$ 0.23 & \cellcolor{indombg} 87.08 $\pm$ 0.39 & 59.30 $\pm$ 3.16\\
\cmidrule(lr){2-6}
& \name{} xs                    & 22M & 81.64 $\pm$ 0.16 & \cellcolor{indombg} 83.31 $\pm$ 0.16 & 47.61 $\pm$ 3.46\\
& \name{} base                    & 86M & \textbf{82.07 $\pm$ 1.28} & \cellcolor{indombg} 86.87 $\pm$ 0.26 & 55.30 $\pm$ 1.49\\
& \name{} large                   & 304M & 81.59 $\pm$ 0.04 & \cellcolor{indombg} 87.74 $\pm$ 0.44 & 61.67 $\pm$ 3.14\\
\midrule 
\multirow{6}{*}{\makecell{Toxic-\\Chat0124}} & ToxicChat-T5-Large $^*$ & 770M & - & - & \cellcolor{indombg} \textbf{82.21}\\
& Prompt Baseline base           & 86M & 72.78 $\pm$ 2.44 & 65.08 $\pm$ 2.13 &\cellcolor{indombg} 76.57 $\pm$ 0.97\\
& Prompt Baseline large          & 304M & 75.83 $\pm$ 1.30 & 66.30 $\pm$ 2.51 &\cellcolor{indombg} 74.51 $\pm$ 1.21\\
\cmidrule(lr){2-6}
& \name{} xs                    & 22M &  75.19 $\pm$ 0.67 & 63.33 $\pm$ 0.64 & \cellcolor{indombg} 57.81 $\pm$ 2.14 \\
& \name{} base                    & 86M & 78.55 $\pm$ 0.44 & 66.70 $\pm$ 1.31 &\cellcolor{indombg} 68.67 $\pm$ 1.97\\
& \name{} large                   & 304M & 78.03 $\pm$ 1.58 & 67.52 $\pm$ 3.72 &\cellcolor{indombg} 78.58 $\pm$ 1.24\\
\bottomrule
\end{tabular}
} 
\vspace{3pt}
\begin{tablenotes}
\tiny
\item $^\dagger$ AEGIS2.0 and WildGuardMix test set results as reported in \citep{ghosh-etal-2025-aegis2}.
\item $^*$ ToxicChat0124 test set results as reported in the Hugging Face dataset card \texttt{lmsys/toxic-chat}.
\item $^\ddagger$ GuardReasoner is trained on a dataset constructed from multiple sources including AEGIS, WildGuardMix, and ToxicChat. Thus, the results are in-domain.
\end{tablenotes}
\end{threeparttable}

\caption{Prompt classification performance of \name{} compared with baseline models, reported using unsafe F1 scores. The results for \name{} are presented as the mean$\pm$standard deviation over three runs with three different random seeds. Gray (\raisebox{.5ex}{\fcolorbox{black}{indombg}{\rule{0pt}{0pt}}}) cells indicate in-domain performance, while white (\raisebox{.5ex}{\fcolorbox{black}{white}{\rule{0pt}{0pt}}}) cells indicate out-of-domain performance.}
\label{tab:PC-results}
\end{table*}

\begin{table*}[t]
\centering
\resizebox{0.74\textwidth}{!}{%
\begin{tabular}{ll@{\hskip 15pt}c@{\hskip 15pt}c@{\hskip 15pt}c@{\hskip 15pt}c}
\toprule
\multirow{2}{*}{\makecell{\textbf{Train}\\\textbf{Dataset}}} & \multirow{2}{*}{\textbf{Model}} & \multirow{2}{*}{\makecell{\textbf{Model}\\\textbf{Size}}} & \multicolumn{3}{c}{\textbf{Test sets}} \\
\cmidrule(lr){4-6} 
&&&\makecell{AEGIS-\\2.0} & \makecell{Wild-\\GuardMix} & \makecell{Toxic-\\Chat0124} \\
\midrule
\multirow{9}{*}{\makecell{AEGIS-\\2.0}} 
&LIME Baseline base & - & \cellcolor{indombg} 31.18 & 21.60 & 6.30\\
&LIME Baseline large & - & \cellcolor{indombg} 31.74 & 22.89 & 9.42\\
\cmidrule(lr){2-6}

&SHAP Baseline base & - & \cellcolor{indombg}  41.07& 25.73 & 9.03\\
&SHAP Baseline large & - & \cellcolor{indombg}  45.94& 32.16& 20.94\\
\cmidrule(lr){2-6} 

& Word Baseline base           & 86M & \cellcolor{indombg} 64.06 $\pm$ 0.78 & 58.33 $\pm$ 0.71 & 50.74 $\pm$ 1.36\\
& Word Baseline large          & 304M & \cellcolor{indombg} 69.53 $\pm$ 0.61 & 61.98 $\pm$ 0.95 & 57.22 $\pm$ 1.43\\
\cmidrule(lr){2-6} 
& \name{} xs                    & 22M & \cellcolor{indombg} 72.73 $\pm$ 0.27 & 53.28 $\pm$ 1.05 & 51.19 $\pm$ 0.39\\
& \name{} base                    & 86M & \cellcolor{indombg} 76.95 $\pm$ 0.54 & 60.40 $\pm$ 0.41 & 59.78 $\pm$ 0.36\\
& \name{} large                   & 304M & \cellcolor{indombg} \textbf{79.60 $\pm$ 0.73} & \textbf{66.66 $\pm$ 0.72} & \textbf{63.18 $\pm$ 0.58}\\
\midrule 
\multirow{9}{*}{\makecell{Wild-\\GuardMix}} 
&LIME Baseline base & - & 32.53 & \cellcolor{indombg} 30.73 & 8.40\\
&LIME Baseline large & - & 32.74 & \cellcolor{indombg} 33.40 & 10.03\\
\cmidrule(lr){2-6} 

&SHAP Baseline base & - & 37.21 & \cellcolor{indombg} 33.44 & 7.28\\
&SHAP Baseline large & - & 43.99 & \cellcolor{indombg} 39.88 & 15.76\\
\cmidrule(lr){2-6} 

& Word Baseline base           & 86M & 66.91 $\pm$ 0.93 & \cellcolor{indombg} 67.24 $\pm$ 0.43 & 50.50 $\pm$ 0.63\\
& Word Baseline large          & 304M & 70.90 $\pm$ 0.53 & \cellcolor{indombg} 70.36 $\pm$ 0.47& 55.45 $\pm$ 2.10\\
\cmidrule(lr){2-6} 
& \name{} xs                    & 22M &69.49 $\pm$ 0.28 & \cellcolor{indombg} 71.17 $\pm$ 0.43 & 48.77 $\pm$ 1.70\\
& \name{} base                    & 86M & 74.28 $\pm$ 0.47 &\cellcolor{indombg} 73.16 $\pm$ 0.52  & 58.86 $\pm$ 0.33\\
& \name{} large                   & 304M &  \textbf{76.93 $\pm$ 0.14} &\cellcolor{indombg} \textbf{75.83 $\pm$ 0.50} & 61.56 $\pm$ 1.06\\
\hline 
\multirow{4}{*}{\makecell{Toxic-\\Chat0124}} 
&LIME Baseline base & - & 29.59 & 20.11 & \cellcolor{indombg} 13.63  \\
&LIME Baseline large & - & 29.34 & 18.19 & \cellcolor{indombg} 17.69 \\
\cmidrule(lr){2-6} 

&SHAP Baseline base & - & 33.12 & 25.46 & \cellcolor{indombg} 11.78  \\
&SHAP Baseline large & - & 40.88 & 30.56 & \cellcolor{indombg}  21.47 \\
\cmidrule(lr){2-6}

& Word Baseline base           & 86M & 45.72 $\pm$ 0.30 & 46.32 $\pm$ 0.34 & \cellcolor{indombg} 38.62 $\pm$ 0.88\\
& Word Baseline large          & 304M & 52.03 $\pm$ 0.95 & 47.89 $\pm$ 2.23 & \cellcolor{indombg} 45.49 $\pm$ 2.01\\
\cmidrule(lr){2-6} 
& \name{} xs                    & 22M & 26.48 $\pm$ 3.74& 23.39 $\pm$ 5.43 & \cellcolor{indombg} 44.63 $\pm$ 3.82\\
& \name{} base                    & 86M & 45.91 $\pm$ 2.31 & 33.77 $\pm$ 5.20 & \cellcolor{indombg} 60.62 $\pm$ 0.20\\
& \name{} large                   & 304M & 52.77 $\pm$ 0.88 & 38.07 $\pm$ 4.41 & \cellcolor{indombg} \textbf{65.99 $\pm$ 0.44}\\
\bottomrule
\end{tabular}
} 
\caption{Explainability classification performance of \name{} compared with baseline models, reported using unsafe F1 scores. The results for \name{} are presented as the mean$\pm$standard deviation over three runs with three different random seeds. Gray (\raisebox{.5ex}{\fcolorbox{black}{indombg}{\rule{0pt}{0pt}}}) cells indicate in-domain performance, while white (\raisebox{.5ex}{\fcolorbox{black}{white}{\rule{0pt}{0pt}}}) cells indicate out-of-domain performance.}
\label{tab:EC-results}
\end{table*}

\begin{table*}[t]
\centering
\resizebox{.6\textwidth}{!}{%
\begin{tabular}{llcccccc}
\toprule

& & \multicolumn{6}{c}{Test Set}\\
\cmidrule(lr){3-8} 
Train set & Ablation & \multicolumn{2}{c}{AEGIS2.0} & \multicolumn{2}{c}{WildGuardMix}  & \multicolumn{2}{c}{ToxicChat0124}\\

\cmidrule(lr){3-8} 
& & base & large & base & large & base & large\\

\midrule
\multirow{4}{*}{\makecell{AEGIS-\\2.0}} 
& Full input &\textbf{86.5} & \textbf{87.75}&\textbf{74.76} & \textbf{78.68}& \textbf{68.16}& \textbf{71.18}\\
& Mask top 1 & 66.67 & 66.11 &69.52 & 72.44 &55.69 & 55.79\\
& Mask top 2 & 52.02& 50.31 & 63.82& 67.15 &48.61 &46.13 \\
& Mask top 3 & 40.44& 37.8 &60.15 & 61.86&41.39 &38.48 \\

 \midrule 
\multirow{4}{*}{\makecell{Wild-\\GuardMix}}
& Full input  &\textbf{83.55} & \textbf{81.56}&\textbf{87.09} &\textbf{87.97} &\textbf{55.73} & \textbf{63.55}\\
& Mask top 1 & 69.76 & 72.29& 83.28 & 83.91& 46.62 & 53.52 \\
& Mask top 2 &59.5 &59.75 &80.51 & 80.15 &42.31 & 47.5\\
& Mask top 3 &46.87 & 47.05&76.99 & 77.68 &37.86 & 45.95 \\
  \midrule 
\multirow{4}{*}{\makecell{Toxic-\\Chat0124}} 
& Full input  &\textbf{78.99} &\textbf{76.38} &\textbf{68.21} & \textbf{63.81}& \textbf{67.43}& \textbf{79.76}\\
& Mask top 1 & 56 & 49.7& 65.21 &52.32 & 63.17& 70.94 \\
& Mask top 2 & 41.37 & 32.25 &62.97 & 41.02 &56.42 & 63.23\\
& Mask top 3 &31.62 & 23.03 &58.8 & 33.73 &37.1 & 56.84\\

\bottomrule
\end{tabular}
} 
\caption{Faithfulness evaluation of the explanations generated by \name{}. In this table, ``base'' refers to \name{} base and ``large'' refers to \name{} large. The reported scores are the unsafe F1 scores for the prompt classification performance of \name{}.}
\label{tab:faithfulness}
\end{table*}

\section{Experimental setup}
\label{sec:exp-setup}
\paragraph{Dataset description:}
\label{sec:dataset-description}
We evaluate \name{} on three prompt safety datasets: AEGIS2.0 \citep{ghosh-etal-2025-aegis2}, WildGuardMix \citep{han2024wildguardopenonestopmoderation}, and ToxicChat0124 \citep{lin-etal-2023-toxicchat}. Each dataset originally provides binary prompt-level safety labels, which we extend with word-level explanation labels using the procedure described in Section~\ref{sec:training-word-label-generation}. Detailed dataset descriptions are included in Appendix~\ref{app:dataset-details}. Appendix~\ref{app:lexical-overlap} further presents a lexical similarity analysis, showing low similarity between all combinations of training and test sets in both in-domain and out-of-domain settings, underscoring the effectiveness of these datasets for evaluating model robustness in both in-domain and out-of-domain scenarios.

\paragraph{Baselines:}
\label{sec:baselines}
We include several LLM-based guardrails as external baselines. Llama Guard \citep{inan2023llamaguardllmbasedinputoutput} is designed for real-time moderation of conversational inputs and outputs. LLAMA3.1 AEGISGUARD \citep{ghosh-etal-2025-aegis2} uses an ensemble of expert classifiers to provide robust online content filtering. ToxicChat-T5-Large \citep{lin-etal-2023-toxicchat} is fine-tuned specifically on adversarial and toxic prompts to improve classification in real-world dialogue systems. WILDGUARD \citep{han2024wildguardopenonestopmoderation} is another recent LLM-based classifier trained on the WildGuardMix dataset. We also report OpenAI Moderation API result. In addition, we report results for two variants of Llama Prompt Guard 2, a recent lightweight guardrail model. In addition to these existing baselines, we develop a set of baselines that mirror the components of \name{} but exclude multi-task learning setup:
{\flushleft \textbf{Prompt Baseline:}} A single-task classifier trained solely to predict whether a prompt is safe or unsafe. It uses the same backbone as \name{} but does not include any explanation mechanism. 
{\flushleft \textbf{Word Baseline:}} A token classifier trained independently to label each word as safe or unsafe. It also uses the same backbone as \name{}. 
{\flushleft \textbf{LIME baseline:}} A post-hoc explanation baseline in which we apply LIME \citep{ribeiro2016lime}  to the Prompt Baseline model to generate word-level explanations after training. 
{\flushleft \textbf{SHAP baseline:}} Another post-hoc explanation method that assigns word-level importance scores using Shapley values \citep{10.5555/3295222.3295230}. The detailed working mechanism of LIME and SHAP baseline is provided in Appendix~\ref{app:posthoc-baseline-details}. We implement two variants of Prompt, Word, LIME, and SHAP baseline: a base version using the DeBERTa-v3-base backbone, and a large version using the DeBERTa-v3-large backbone.
\paragraph{Our models (\name{}):} For our experiments, we implement three versions of \name{}: \textit{xs}, \textit{base}, and \textit{large}. All three share the same architecture described in Section~\ref{sec:proposed-method}, but differ in the choice of encoder. \name{} xs uses DeBERTa-v3-xsmall (22M parameters), \name{} base uses DeBERTa-v3-base (86M parameters), and \name{} large uses DeBERTa-v3-large (304M parameters).
\paragraph{Hyperparameters:} We train all models using the following hyperparameters: a learning rate of $2 \times 10^{-5}$, batch size of 16, and 3 training epochs. The optimizer is AdamW. Each experiment is repeated with three random seeds (42, 52, and 62).
\section{Analysis and discussion}
In this section, we compare the performance of \name{} with other baselines. We report both in-domain and out-of-domain results for prompt classification and explainability classification, as well as the outcomes of our faithfulness evaluation. All reported scores are F1 scores for the unsafe class. For \name{} and our custom baselines, we run three experiments with different random seeds and report the mean$\pm$standard deviation. For in-domain performance, we train \name{} base and \name{} large on the AEGIS2.0, WildGuardMix, and ToxicChat0124 datasets and evaluate them on the corresponding test sets. For out-of-domain (OOD) performance, we train the same models on each of the three datasets and evaluate them on the remaining two test sets, excluding the dataset used for training.

\subsection{Prompt classification performance}
\paragraph{In-domain performance:} The gray cells (\raisebox{.5ex}{\fcolorbox{black}{indombg}{\rule{0pt}{0pt}}}) in Table~\ref{tab:PC-results} indicate in-domain prompt classification performance. On the AEGIS2.0 test set, \name{} large achieves the highest F1 score of 87.54\%, outperforming other models. On the WildGuardMix dataset, the best model is WILDGUARD with an F1 score of 88.9\%. However, \name{} base (87.09\%) and \name{} large (87.97\%) achieve comparable results despite being significantly smaller. This demonstrates that although WILDGUARD is built on an 8B parameter LLM, our models with 86M and 304M parameters achieve similar performance. On the ToxicChat0124 dataset, the best-performing model is ToxicChat-T5-Large, but \name{} large delivers comparable performance despite its smaller size. In contrast, \name{} base underperforms on ToxicChat0124 relative to other models. We further analyze this issue in the error analysis section (Appendix \ref{app:error-analysis}) and show that performance can be improved with additional tuning. Overall, the in-domain results demonstrate that \name{}, while considerably smaller, delivers strong performance that rivals or exceeds much larger models.
\paragraph{Out-of-domain performance:} The white cells (\raisebox{.5ex}{\fcolorbox{black}{white}{\rule{0pt}{0pt}}}) in Table~\ref{tab:PC-results} indicate out-of-domain (OOD) prompt classification performance. On the AEGIS2.0 test set, \name{} large trained on WildGuardMix achieves the highest F1 score of 82.07\%, outperforming all other models. On the WildGuardMix test set, LLAMA3.1 AEGISGUARD achieves the best result. However, \name{} large trained on AEGIS2.0 delivers a comparable score despite being an order of magnitude smaller (304M vs. 8B parameters). On the ToxicChat0124 test set, \name{} large trained on AEGIS2.0 achieves the best performance at 69.98\%, substantially outperforming the 2024 OpenAI Moderation API (61.41\%). In addition to \name{} base and \name{} large, we present results for a super lightweight version, \name{} xs, with only 22M parameters. Two similar lightweight models (Llama Prompt Guard 2) were recently released by Meta, motivating us to develop \name{} xs for comparison. As shown in Table~\ref{tab:PC-results}, despite its small size, \name{} xs performs strongly, whereas variations of Llama Prompt Guard 2 perform poorly. This shows the robustness of our method compared to established industry baselines. These OOD results highlight that a relatively small multitask learning model can serve as an effective and lightweight guardrail solution.

\subsection{Explainability classification performance}

\paragraph{In-domain performance:} 
The gray cells (\raisebox{.5ex}{\fcolorbox{black}{indombg}{\rule{0pt}{0pt}}}) in Table~\ref{tab:EC-results} indicate in-domain explanation classification performance. Across all datasets, \name{} base and \name{} large consistently outperform the baseline models. \name{} large achieves the best results with 79.60\% on AEGIS2.0, 75.83\% on WildGuardMix, and 65.99\% on ToxicChat0124, followed by \name{} base. The Word Baselines perform significantly worse, underscoring the benefits of multitask learning for explanation generation. Moreover, LIME and SHAP baselines underperform, suggesting that post hoc explanation methods are less effective, whereas our multitask learning approach produces stronger and more reliable results.

\paragraph{Out-of-domain performance:} 
The white cells (\raisebox{.5ex}{\fcolorbox{black}{white}{\rule{0pt}{0pt}}}) in Table \ref{tab:EC-results} indicate OOD explainability classification performance. On the AEGIS2.0 test set, \name{} large trained on WildGuardMix achieves the highest F1 score of 76.93\%. On the WildGuardMix test set, \name{} large trained on AEGIS2.0 outperforms all baselines. On the ToxicChat0124 test set, \name{} large trained on AEGIS2.0 again leads with 63.18\%. Across all three datasets, \name{} consistently outperforms Word, LIME, and SHAP baseline, reinforcing the effectiveness of multitask learning for explainability classification.

\begin{table*}[t]
\centering
\resizebox{\textwidth}{!}{%
\begin{tabular}{llcccccccc cccc}
\toprule

& & \multicolumn{12}{c}{Test Set}\\
\cmidrule(lr){3-14} 
Train set & Ablation & \multicolumn{4}{c}{AEGIS2.0} & \multicolumn{4}{c}{WildGuardMix}  & \multicolumn{4}{c}{Toxic-Chat0124}\\
\cmidrule(lr){3-14} 
& & \multicolumn{2}{c}{base} & \multicolumn{2}{c}{large} & \multicolumn{2}{c}{base} & \multicolumn{2}{c}{large} & \multicolumn{2}{c}{base} & \multicolumn{2}{c}{large}\\
\cmidrule(lr){3-14} 
& & PC & EC & PC & EC & PC & EC & PC & EC & PC & EC & PC & EC\\
\midrule
\multirow{2}{*}{AEGIS2.0} & Full joint loss (all terms) & \textbf{86.50}&76.38&\textbf{87.75}&79.71&74.76&\textbf{60.84}&78.68&\textbf{67.46}&\textbf{68.16}&\textbf{59.46}&\textbf{71.18}&\textbf{63.69}\\

&Remove $\delta_p$, $\delta_t$ and focal loss & 86.45&\textbf{77.3}&87.62&\textbf{79.97}&\textbf{74.8}&59.02&\textbf{79.60}&66.02&67.67&58.79&68.44&62.72\\

 \midrule 
\multirow{2}{*}{WildGuardMix} &Full joint loss (all terms) & \textbf{83.55}&\textbf{74.75}&\textbf{81.56}&76.83&\textbf{87.09}&72.57&87.97&76&55.73&58.61&\textbf{63.55}&\textbf{61.86}\\

&Remove $\delta_p$, $\delta_t$ and focal loss & 80.74&73.7&81.14&\textbf{76.84}&86.43&\textbf{73.42}&\textbf{88.48}&\textbf{76.1}&\textbf{56.48}&\textbf{59.66}&59.62&59.87\\

  \midrule 
\multirow{2}{*}{ToxicChat0124} &Full joint loss (all terms) & \textbf{78.99}&48.15&\textbf{76.38}&\textbf{51.99}&\textbf{68.21}&\textbf{37.83}&\textbf{63.81}&33.4&67.43&60.85&\textbf{79.76}&\textbf{66.19}\\

&Remove $\delta_p$, $\delta_t$ and focal loss & 78.42&\textbf{48.61}&73.83&51.19&65.92&29.78&63.45&\textbf{36.18}&\textbf{70.69}&\textbf{62.79}&78.42&65.87\\

\bottomrule
\end{tabular}
} 
\caption{Ablation results for the joint loss function. ``base'' and ``large'' denote \name{}-base and \name{}-large, respectively. ``PC'' and ``EC'' correspond to the prompt classifier and explainability classifier. All values report F1 scores.}
\label{tab:ablation-result}
\end{table*}

\subsection{Faithfulness evaluation}
To assess the faithfulness of the explanations generated by \name{}, we adopt a word-masking perturbation test. The experimental procedure is as follows: First, we use \name{} to predict word labels for the input prompt. Next, we rank the words predicted as unsafe by their classifier confidence scores (predicted probabilities). We then mask the top-k words and re-evaluate the prompt classification performance of \name{} on the modified input. This procedure directly tests whether the words highlighted as unsafe are indeed causally important to the model’s decision. As shown in Table~\ref{tab:faithfulness}, masking the top-1, top-2, and top-3 predicted unsafe words consistently degrades the prompt classification performance of \name{}, with larger drops observed as more tokens are masked. These results indicate that the \name{}'s prompt classifier relies on the highlighted explanation words when making its decisions, confirming that the generated explanations are faithful.

\subsection{Ablation study of the joint loss function}
\label{app:ablation_study}

We introduce a novel auxiliary supervision mechanism in the joint loss function (Equation~\ref{eq:joint-loss}) through the weak supervision signals $\delta_p$ and $\delta_t$, together with focal loss modulation. In this ablation study, we evaluate whether these components provide a meaningful contribution to model performance. We compare the following two settings:

\begin{enumerate}[label=\alph*), itemsep=0pt]
    \item \textbf{Full joint loss (all terms):} The complete loss function, including $\delta_p$, $\delta_t$, and focal loss modulation.
    \item \textbf{Without weak supervision and focal loss:} Both weak supervision signals ($\delta_p$, $\delta_t$) and focal loss are removed, leaving only the standard cross-entropy loss for prompt classification $\mathcal{L}_{\text{pc}}$ and token classification $\mathcal{L}_{\text{ec}}$.
\end{enumerate}

Table~\ref{tab:ablation-result} reports results across both in-domain and out-of-domain evaluations, covering a total of 36 experimental settings. Overall, the full joint loss outperforms the ablated variant in 22 out of 36 cases. Notably, among the 24 out-of-domain evaluations, the full joint loss achieves better performance in 17 cases, corresponding to an improvement in approximately 71\% of out-of-domain scenarios. This consistent advantage in out-of-domain settings indicates that the proposed weak supervision mechanism improves robustness and generalization. Taken together, these results demonstrate that the weak supervision signals $\delta_p$ and $\delta_t$, along with their focal-loss-based modulation, make a positive contribution to model performance. Notably, this improvement is achieved with only a constant-time overhead added to the training loop.

\subsection{Other evaluations}
Appendix~\ref{app:computational-efficiency} analyzes computational efficiency, demonstrating that \name{} achieves lower inference time and GPU memory usage than other guardrail models. Appendix~\ref{app:generalization-of-explainability} shows that although the explainability classifier is trained using synthetic labels from \texttt{GPT-4o-mini}, its performance is not upper-bounded by it. Finally, Appendices~\ref{app:fine-grained-performance} and~\ref{app:unseen-category-performance} show that \name{} maintains strong performance across fine-grained risk categories and remains robust to previously unseen categories. Appendix~\ref{app:eval-benign-prompt} evaluates robustness to exaggerated safety on benign prompts with unsafe vocabulary and shows that \name{} remains robust.
\section{Conclusion}
We introduced a lightweight explainable guardrail, which jointly classifies prompts as safe or unsafe and explains its decision by highlighting words that contribute to the prediction. Despite being considerably smaller than state-of-the-art approaches, our method achieves competitive or superior performance in both in-domain and out-of-domain settings compared to existing guardrail methods.

\section*{Limitations}
Our evaluation is conducted primarily on English-language datasets. Although the ToxicChat0124 dataset contains a small number of multilingual prompts, this subset is insufficient to meaningfully assess performance in multilingual settings. While our method is designed to be language-agnostic in principle and should generalize to other languages when appropriate training data from those domains are available, we do not explicitly evaluate multilingual performance in this work. A systematic evaluation across multiple languages and domains is left for future work.

\section*{Acknowledgments}
This work was partially supported by NVIDIA Corporation through the NVIDIA Academic Grant Program, which provided a DGX Spark machine that supported part of the computational and GPU requirements of this work.
Mihai Surdeanu declares a financial interest in lum.ai. This interest has been properly disclosed to the University of Arizona Institutional Review Committee and is
managed in accordance with its conflict of interest policies.

\bibliography{anthology,custom}

\appendix

\section{\name{} architecture details}
\label{app:model-architecture}
This appendix expands Section \ref{sec:architecture} by providing a detailed description of each component of \name{}.
\subsection{Shared encoder}
We use a shared transformer encoder as the backbone for both the prompt classification and explanation generation tasks. 
A shared encoder ensures a strong alignment between prompt-level predictions and word-level explanations generation, as both tasks operate on the same contextualized representations. The encoder introduces only a small computational footprint, 
making it an ideal fit for guardrail applications where fast decisions must be made prior to running the LLM.

\subsection{Attention pooling layer}
The attention pooling layer computes a fixed-length vector from the hidden states of the encoder to serve as input to the prompt classifier. We adopt a simplified version of the attention pooling method proposed by \citet{yang-etal-2016-hierarchical}, using a single-layer linear transformation without nonlinearity. Let \( H = [h_0, h_1, \dots, h_T] \in \mathbb{R}^{T \times d} \) be the sequence of hidden states. The pooled representation \( v \in \mathbb{R}^d \) is computed as:
\begingroup
\setlength{\abovedisplayskip}{1pt}
\setlength{\belowdisplayskip}{2pt}
\[
\alpha_i = \text{softmax}(w^\top h_i + b), \quad v = \sum_{i=1}^{T} \alpha_i h_i
\]
\endgroup
where \( w \in \mathbb{R}^d \) and \( b \in \mathbb{R} \) are learnable parameters.

\subsection{Prompt classifier}
The prompt classifier predicts whether a given prompt is safe or unsafe based on the fixed-length representation \( v \in \mathbb{R}^d \) obtained from the attention pooling layer. This representation is passed through a linear classifier to produce the final prediction:
\begingroup
\setlength{\abovedisplayskip}{3pt}
\setlength{\belowdisplayskip}{3pt}
\[
\hat{y}_{\text{prompt}} = \text{softmax}(W_p v + b_p)
\]
\endgroup
where \( W_p \in \mathbb{R}^{2 \times d} \) and \( b_p \in \mathbb{R}^2 \) are learnable parameters.

\subsection{Explainability classifier}

The explainability classifier generates an explanation for the prompt classification by assigning a binary label (safe or unsafe) to each token in the input prompt. It operates directly on the contextualized token embeddings \( H = [h_1, h_2, \dots, h_T] \in \mathbb{R}^{T \times d} \) produced by the shared transformer encoder. Each token embedding is passed through a linear classifier to produce the final prediction:
\begingroup
\setlength{\abovedisplayskip}{3pt}
\setlength{\belowdisplayskip}{3pt}
\[
\hat{y}_{\text{token}, i} = \text{softmax}(W_t h_i + b_t)
\]
\endgroup
where \( W_t \in \mathbb{R}^{2 \times d} \) and \( b_t \in \mathbb{R}^2 \) are learnable parameters.

\section{Explanation training label generation procedure}
\label{appendix:explanation-label-generation-detail}
This appendix expands Section \ref{sec:training-word-label-generation}
that discusses  a procedure to generate word labels using an LLM.
\begin{figure}[t]
\rule{\linewidth}{0.5pt}
\textbf{Prompt:}\\
\texttt{``Why is the following prompt considered \textcolor{red}{\{bias term\}}?\\
Provide a list of words or phrases that made you believe the prompt is \textcolor{red}{\{bias term\}} in the following JSON format.
}
\begin{lstlisting}[
  basicstyle=\ttfamily\small,
  showstringspaces=false,
  breaklines=true,
  frame=none,
  backgroundcolor=\color{gray!10}
]
{ 
    "unsafe":True/False, 
    "safe":True/False, 
    "unsafe_keywords":[], 
    "safe_keywords":[], 
    "justification": ""
} 
\end{lstlisting}
\texttt{In the JSON format, "unsafe" or "safe" can be True or False. Depending on your decision, "unsafe\_keywords" or "safe\_keywords" can be empty. Justification should be very short. Don't generate anything except the JSON format.\\
Prompt: \textcolor{red}{\{concatenate prompt here\}}} \\

\rule{\linewidth}{0.5pt}
\caption{Prompt for synthetic word label generation. The bias term is set to \texttt{safe} for Query~1 and \texttt{unsafe} for Query~2, as described in Section~\ref{sec:training-word-label-generation}.}
\label{fig:gpt-prompt}
\end{figure}
\subsection{Prompt for extracting word labels}
In this section, we describe the prompt design used to generate word level labels for our dataset using \texttt{GPT-4o-mini}. The full prompt is shown in Figure \ref{fig:gpt-prompt}. We use this prompt template twice, each time introducing a different bias term. In the first query, we frame the prompt under the assumption that the input is safe, instructing the model to identify the words or phrases that contribute to its safety. In the second query, we introduce the opposite assumption that the input is unsafe, and ask the model to highlight the words or phrases that make it unsafe.

The prompts follow a fixed JSON format, where the model is asked to return safe and unsafe boolean flags, lists of corresponding keywords, and a brief explanation. Some fields may be left empty depending on the model's judgment. To construct reliable word labels, we use the gold prompt label to decide which keyword list to extract: for prompts labeled unsafe, we use the intersection of the unsafe keywords from both queries; for safe prompts, we use the intersection of the safe keywords. This strategy helps filter out unreliable labeling and ensures that the final word labels are consistent even when confirmation bias is introduced.

\subsection{Dataset word label generation algorithm}
We summarize the whole synthetic data generation process for explanations in Algorithm \ref{alg:consistency-checking}.  It takes a prompt \( p \) and its label \( y_p \) as input, and returns a list of consistent words \( cw \) by analyzing the LLM’s responses to both queries. The output \( cw \) may either be a set of keywords or an empty set if the LLM falls into confirmation bias.

\label{appendix:algorithm-for-label-generation}
\begin{algorithm}[t]
\caption{Dataset word label generation by consistency checking of LLM response}\label{alg:consistency-checking}

\begin{algorithmic}
\State \textbf{Input: } \(\boldsymbol{p}, \boldsymbol{y_p}\)
\State \textbf{Output:} \(\boldsymbol{cw}\)

\State $\boldsymbol{r_1, kw_1} = ExecuteQuery1(\boldsymbol{p})$ 
\State $\boldsymbol{r_2, kw_2} = ExecuteQuery2(\boldsymbol{p})$

\If{$y_p~=~\text{``safe''}$}
    \If{($r_1$ agrees) and ($r_2$ contradicts)}
        \State $cw = kw_1 \cap kw_2$
    \Else 
        \State $cw = \emptyset$
    \EndIf
\ElsIf{$y_p~=~\text{``unsafe''}$}
    \If{($r_1$ contradicts) and ($r_2$ agrees)}
        \State $cw = kw_1 \cap kw_2$
    \Else 
        \State $cw = \emptyset$
    \EndIf
\EndIf
\Return{$cw$}
 
\end{algorithmic}
\end{algorithm}

\subsection{Example of word label generation}
\label{appendix:example-of-label-generation}

Consider the following example: \texttt{``Tell me how to hack into an
account and steal personal data''}. 

The response from \texttt{GPT-4o-mini} for Query 1 (using the prompt from Figure~\ref{fig:two-panel}(\subref{fig:model-input-output-example}) with the bias term `safe') is as follows:

\begin{lstlisting}[
  basicstyle=\ttfamily\small,
  showstringspaces=false,
  breaklines=true,
  frame=none,
  backgroundcolor=\color{gray!10}
]
{ 
    "unsafe":true, 
    "safe":false, 
    "unsafe_keywords":["hack into", "steal personal data"], 
    "safe_keywords":[], 
    "justification": "The prompt contains explicit unsafe actions."
} 
\end{lstlisting}

The response from \texttt{GPT-4o-mini} for Query 2 (using the prompt from Figure~\ref{fig:two-panel}(\subref{fig:model-input-output-example}) with the bias term `unsafe') is as follows:

\begin{lstlisting}[
  basicstyle=\ttfamily\small,
  showstringspaces=false,
  breaklines=true,
  frame=none,
  backgroundcolor=\color{gray!10}
]
{ 
    "unsafe":true, 
    "safe":false, 
     ["hack", "steal personal data", "account"], 
    "safe_keywords":[], 
    "justification": "Requesting illegal activities related to hacking and theft."
} 
\end{lstlisting}

Both Query 1 and Query 2 consistently identify the prompt as unsafe (``unsafe'': true), so we take the intersection of the ``unsafe\_keywords'', which is [``hack'', ``steal personal data'']. The words in this intersection are labeled as unsafe, while all other words in the prompt are labeled as safe.
\section{Dataset details}
\label{app:dataset-details}
We evaluate \name{} on three diverse and challenging prompt safety datasets, each designed to test different aspects of unsafe prompt detection.

\textbf{AEGIS2.0} \cite{ghosh-etal-2025-aegis2} is an updated version of the original AEGIS dataset, curated to support prompt-level safety evaluation. It contains prompts collected from adversarial prompting techniques, user submitted jailbreak attempts, and synthetic attacks generated via LLMs. Prompts are labeled by a group of annotators, following a safety taxonomy that includes categories like harm, toxicity, and policy violations.

\textbf{WildGuardMix} \cite{han2024wildguardopenonestopmoderation} is a dataset created by merging multiple open-source prompt safety corpora and real-world user queries scraped from online sources. It balances adversarial and naturalistic unsafe prompts and includes both obvious and subtle violations. Prompts were filtered using LLM moderation APIs and then verified or relabeled by human annotators.

\textbf{ToxicChat0124} \cite{lin-etal-2023-toxicchat} comprises real-world user prompts collected from chatbot logs and publicly shared datasets with consent. It emphasizes subtle, context-dependent toxicity and is highly imbalanced with fewer than 7\% of prompts are labeled as unsafe. Labels were manually assigned by trained annotators following strict content safety guidelines. We use the 2024 version of this dataset.

While each dataset originally includes binary safety labels at the prompt level, we extend them with word level explanation labels using the procedure described in section \ref{sec:training-word-label-generation}. Using this approach, we were able to generate word labels for 65.7\% of the instances in AEGIS2.0, 66.7\% in WildGuardMix, and 85.8\% in ToxicChat0124.

\begin{table*}[h]
\centering
\small
\resizebox{.7\textwidth}{!}{%
\begin{tabular}{c cc cc cc}
\hline
\textbf{\makecell{Word \\ Supervision (\%)}} & \multicolumn{2}{c}{\textbf{AEGIS 2.0}} & \multicolumn{2}{c}{\textbf{WildGuardMix}} & \multicolumn{2}{c}{\textbf{ToxicChat0124}} \\
 & \textbf{PC} & \textbf{EC} & \textbf{PC} & \textbf{EC} & \textbf{PC} & \textbf{EC} \\
\hline
20      & 86.86 & 73.94 & 86.22 & 71.82 & 68.23 & 54.38 \\
40      & 86.83 & 75.15 & 86.48 & 72.51 & 68.10 & 57.42 \\
60      & 87.04 & 76.54 & 87.01 & 73.05 & 70.88 & 58.78 \\
65--80  & 86.63 & 76.64 & 86.40 & 73.25 & 69.98 & 59.05 \\
86      &        &        &        &        & 70.37 & 59.91 \\
\hline
\end{tabular}
}
\caption{Performance of \name{} base under varying levels of explanation label coverage. PC denotes prompt classification performance and EC denotes explanation classification performance.}
\label{tab:word_supervision_ablation}
\end{table*}

\section{Impact of partial explanation label coverage}
\label{app:partial-explanation-coverage}
The synthetic explanation labeling method described in Section~\ref{sec:training-word-label-generation} relies on confirmation bias to ensure high-quality word-level supervision. Using this approach, we are able to generate explanation labels for 65.7\% of instances in AEGIS2.0, 66.7\% in WildGuardMix, and 85.8\% in ToxicChat0124. For instances without explanation labels, we train the model using only prompt-level supervision by setting the explanation loss to zero. This is naturally supported by our multi-task learning framework, where the shared encoder learns from both tasks, allowing the model to benefit from explanation supervision when available and rely on prompt supervision otherwise.

To analyze the effect of partial explanation supervision, we conducted an ablation study where we trained \name{} base using randomly selected subsets of 20\%, 40\%, 60\%, and the full available (65--86\%) explanation labels, discarding the rest during training. The results in Table~\ref{tab:word_supervision_ablation} show that \name{} base is able to learn meaningful explanation patterns even with only 20\% supervision. As the coverage increases, performance improves, but the gains diminish beyond 40\%. In most cases, increasing supervision beyond this point results in less than 2\% improvement in F1 score, indicating that performance largely stabilizes.

Overall, these results demonstrate that partial explanation coverage does not significantly hinder learning. Approximately 40\% explanation supervision is sufficient to achieve strong explanation classification performance, highlighting the robustness of our approach under limited but high-quality annotation availability.
\begin{table*}[h]
\centering
\resizebox{\textwidth}{!}{%
\setlength{\tabcolsep}{4pt}
\begin{tabular}{llcccccccccc}
\toprule
\textbf{Train set} & \textbf{Test set} & \textbf{[0-0.1)} & \textbf{[0.1-0.2)} & \textbf{[0.2-0.3)} & \textbf{[0.3-0.4)} & \textbf{[0.4-0.5)} & \textbf{[0.5-0.6)} & \textbf{[0.6-0.7)} & \textbf{[0.7-0.8)} & \textbf{[0.8-0.9)} & \textbf{[0.9-1.0)} \\
\midrule
AEGIS & WildGuard & 43.3 & 31.8 & 17.6 & 4.9 & 1.7 & 0.4 & 0.2 & 0.0 & 0.0 & 0.0 \\
AEGIS & ToxicChat & 12.3 & 38.2 & 28.3 & 9.4 & 3.0 & 4.9 & 1.1 & 0.3 & 0.1 & 2.4 \\
\hline
WildGuard & AEGIS & 4.0 & 24.6 & 29.7 & 15.2 & 5.7 & 7.6 & 2.2 & 0.5 & 0.2 & 10.2 \\
WildGuard & ToxicChat & 11.5 & 37.3 & 30.1 & 10.8 & 2.7 & 4.5 & 1.0 & 0.2 & 0.2 & 1.6 \\
\hline
ToxicChat & AEGIS & 16.9 & 41.9 & 29.8 & 6.5 & 2.2 & 2.1 & 0.4 & 0.0 & 0.1 & 0.2 \\
ToxicChat & WildGuard & 55.4 & 32.6 & 11.2 & 0.7 & 0.1 & 0.0 & 0.1 & 0.0 & 0.0 & 0.0 \\
\bottomrule
\end{tabular}
}
\caption{Unigram-based lexical similarity distribution between out-of-domain training and test sets. Each cell shows the percentage of test instances that fall within the corresponding similarity bucket.}
\label{tab:unigram-similarity-distribution}

\end{table*}

\begin{table*}[h]
\centering
\resizebox{\textwidth}{!}{%
\setlength{\tabcolsep}{2pt}
\begin{tabular}{llcccccccccc}
\toprule
\textbf{Train set} & \textbf{Test set} & \textbf{[0-0.1)} & \textbf{[0.1-0.2)} & \textbf{[0.2-0.3)} & \textbf{[0.3-0.4)} & \textbf{[0.4-0.5)} & \textbf{[0.5-0.6)} & \textbf{[0.6-0.7)} & \textbf{[0.7-0.8)} & \textbf{[0.8-0.9)} & \textbf{[0.9-1.0)} \\
\midrule
AEGIS & WildGuard & 63.5 & 22.6 & 11.2 & 2.2 & 0.4 & 0.1 & 0.0 & 0.0 & 0.0 & 0.0 \\
AEGIS & ToxicChat & 43.1 & 29.2 & 15.2 & 5.0 & 3.0 & 1.9 & 0.8 & 0.5 & 0.2 & 1.1 \\
\hline
WildGuard & AEGIS & 28.1 & 30.2 & 18.4 & 7.2 & 3.3 & 2.7 & 1.0 & 0.2 & 0.1 & 8.8 \\
WildGuard & ToxicChat & 43.0 & 29.8 & 15.3 & 5.1 & 3.0 & 2.2 & 0.6 & 0.3 & 0.1 & 0.5 \\
\hline
ToxicChat & AEGIS & 51.3 & 30.4 & 12.2 & 3.3 & 1.4 & 1.1 & 0.2 & 0.1 & 0.0 & 0.1 \\
ToxicChat & WildGuard & 71.8 & 21.6 & 6.0 & 0.3 & 0.1 & 0.1 & 0.1 & 0.0 & 0.0 & 0.0 \\
\bottomrule
\end{tabular}
}
\caption{Bigram-based lexical similarity distribution between out-of-domain training and test sets. Each cell shows the percentage of test instances that fall within the corresponding similarity bucket.}
\label{tab:bigram-similarity-distribution}
\end{table*}

\section{Lexical overlap between train and test sets}
\label{app:lexical-overlap}

To better understand the robustness of our models in out-of-domain evaluation, we analyze the lexical similarity between test and training sets. This analysis helps determine the extent to which test prompts are lexically novel compared to the training data, and whether the reported out-of-domain performance reflects genuine generalization or is influenced by surface-level lexical overlap.

Specifically, we compute the \textit{maximum Jaccard similarity} between each test prompt and all training prompts. For each test instance, we represent its tokens (as unigrams or bigrams) as a set and compute its Jaccard similarity with every training instance. The highest similarity value is retained as its max Jaccard score. This is formally defined as:

\[
\text{Similarity}(\text{test}_i) = \max_{j \in [1, N_{\text{train}}]} \; \text{Jaccard}(\text{test}_i, \text{train}_j)
\]

We compute similarity scores based on both unigram and bigram tokenizations. For the unigram analysis, we remove common stop words to focus on meaningful content words. In bigram analysis, we keep all tokens without any filtering.

Table~\ref{tab:unigram-similarity-distribution} presents the lexical similarity distribution computed using unigram tokenization, while Table~\ref{tab:bigram-similarity-distribution} shows the results for bigram tokenization.

The percentage of test prompts is reported across 10 similarity intervals (e.g., \([0 \leq s < 0.1),\ [0.1 \leq s < 0.2),\ \dots,\ [0.9 \leq s \leq 1.0]\)). For example, in Table~\ref{tab:unigram-similarity-distribution}, 43.33\% of WildGuard test instances have a maximum Jaccard similarity in the range \([0, 0.1)\) when compared to the AEGIS training set.

For interpretation, we categorize the similarity scores as follows:
\begin{itemize}
    \item \textbf{Low similarity:} \(0 \leq s < 0.3\)
    \item \textbf{Moderate similarity:} \(0.3 \leq s < 0.7\)
    \item \textbf{High similarity:} \(0.7 \leq s \leq 1.0\)
\end{itemize}

Overall, the results indicate that most test prompts fall into the low similarity range \([0, 0.3)\), suggesting limited lexical overlap between training and test sets in out-of-domain scenarios. As expected, the similarity scores are even lower in the bigram setting.

\section{Post-hoc explainability baseline details}
\label{app:posthoc-baseline-details}

\subsection{LIME baseline details}
We follow a procedure similar to that proposed in the original LIME paper \citep{ribeiro2016lime}. This baseline generates explanations through the following steps:
\begin{enumerate}
\item For each input prompt, we run LIME with $N$ perturbed samples created by randomly removing subsets of words. In our experiments, we set $N=1500$ and use bag-of-words features.
\item Each perturbed prompt is passed through the \textit{Prompt Baseline} model to obtain the predicted probability of the target class (``unsafe'').
\item LIME fits a local surrogate model using the perturbed samples and their predicted probabilities, weighted by their similarity to the original prompt. The surrogate is trained using the top-$K$ most informative words, with $K=25$ in our experiments, and produces a weight for each word indicating its contribution toward the ``unsafe'' class.
\item We convert LIME’s word weights into binary labels by tuning a threshold on the dev set to maximize F1.
\item The dev-selected threshold is then applied at test time to obtain word-level safe/unsafe predictions.
\end{enumerate}

\subsection{SHAP baseline details}
We follow a standard SHAP-based post-hoc explanation procedure for text classification \citep{10.5555/3295222.3295230}. This baseline generates explanations through the following steps:
\begin{enumerate}
\item We treat the trained \textit{Prompt Baseline} as a black-box prediction function that maps an input prompt to class probabilities (safe, unsafe).
\item SHAP constructs explanations by systematically masking subsets of input tokens and measuring the change in the predicted probability of the target class (``unsafe''). In contrast to random perturbations, SHAP uses a structured masking strategy that approximates Shapley values, ensuring fair attribution of importance across tokens.
\item For each input prompt, SHAP computes attribution scores for individual tokens that quantify their contribution to the model’s prediction relative to a baseline input.
\item Since SHAP operates at the subword-token level, we aggregate subword attribution scores into word-level scores by summing the contributions of all subword tokens whose character spans overlap with each word.
\item To obtain binary word labels, we threshold the resulting word-level SHAP scores. The threshold is tuned on the dev split of the training data to maximize word-level F1 score, and the same threshold is applied during test-time evaluation. Words with scores above the threshold are labeled as unsafe, and all others are labeled as safe.
\end{enumerate}

\begin{table*}[t]
\centering
\begin{tabular}{lccc}
\toprule
\textbf{Model} & \textbf{Model Size} & \textbf{Inference time (ms/input)} & \textbf{GPU memory use (GB)} \\
\midrule
LEG xs                & 22M   & 7.81               & 1.01 \\
LEG base              & 86M  & 8.28               & 1.67 \\
LEG large             & 304M  & 14.57              & 3.06 \\
Llama Prompt Guard 2  & 22M   & 9.17               & 1.04 \\
Llama Prompt Guard 2  & 86M  & 9.47               & 1.90 \\
DuoGuard              & 500M  & 16.47              & --   \\
Toxic-Chat-T5 Large   & 770M  & 31.95              & 3.68 \\
GuardReasoner         & 1B--8B & 26.66--35.77       & 78.00 \\
Llama Guard 3         & 1B    & 58.88              & --   \\
ShieldGemma           & 2B    & 57.87              & --   \\
\bottomrule
\end{tabular}
\caption{Inference time and GPU memory usage across models.}
\label{tab:inference-time}
\end{table*}

\section{Computational efficiency}
\label{app:computational-efficiency}

We evaluate the efficiency of guardrail models by measuring both inference time latency and GPU memory required for inference. All experiments were performed using an NVIDIA H100 GPU. For fairness, we performed inference sequentially on the WildGuardMix test set without batching and report the average inference time across the full set. Results for DuoGuard, Llama Guard 3, and ShieldGemma were taken from \citep{deng2025duoguardtwoplayerrldrivenframework}, and results for GuardReasoner were taken from \citep{liu2025guardreasonerreasoningbasedllmsafeguards}. Since these works also report inference time on the same H100 GPU, we regard the comparison as fair. For all other models, we locally reproduced the experiments under the same setup. Table~\ref{tab:inference-time} summarizes the results.

The \name{} family is consistently efficient: \name{} xs achieves 7.81 ms per input using only 1.01 GB of memory, while \name{} base and \name{} large remain lightweight at 8.28 ms / 1.67 GB and 14.57 ms / 3.06 GB, respectively. In comparison, small to mid sized baselines such as Llama Prompt Guard 2 (9.17 to 9.47 ms, 1.04 to 1.90 GB), DuoGuard (16.47 ms), and Toxic-Chat-T5 Large (31.95 ms, 3.68 GB) show slower inference and higher memory use. Larger guardrails are substantially more expensive: GuardReasoner requires 26 to 36 ms per input and up to 78 GB of memory, while Llama Guard 3 and ShieldGemma exceed 57 ms.  
Overall, \name{} offers substantial efficiency gains. Compared to GuardReasoner, \name{} xs is over 3$\times$ faster and about 75$\times$ more memory efficient. Similarly, compared to Llama Guard 3 and ShieldGemma, \name{} xs is over 7$\times$ faster. These results show that \name{} achieves significant speedups and memory savings while remaining lightweight across all configurations.

Crucially, \name{} achieves this efficiency while supporting both prompt classification and explanation generation. Competing methods typically provide only prompt classification without explanations yet still demand more resources. This makes \name{} the first guardrail to combine lightweight inference with faithful explanation generation, enabling both efficiency and transparency for real time deployment.
\section{Error analysis}
\label{app:error-analysis}
The in-domain performance of \name{} base on the ToxicChat0124 dataset is lower than that of other baselines. We found this is due to the model exhibiting high recall (89.5\%) but low precision (54.09\%). It is well known that precision can be improved through probability threshold tuning. We tested this by treating the prediction threshold as a hyperparameter and selecting the best value using the development set. With this adjusted threshold, \name{} base achieves an F1 score of 75\% on the in domain ToxicChat0124 dataset. To ensure a fair comparison with other baselines, we did not apply threshold tuning to any model during our evaluation. However, we observed that this tuning strategy improves the performance of almost every variant of \name{}.
\begin{table*}[t]
\centering
\small
\begin{tabular}{l c c c c}
\hline
\textbf{Model} & \textbf{Explanation training data} & \textbf{AEGIS 2.0} & \textbf{WildGuardMix} & \textbf{ToxicChat0124} \\
\hline
LEG Base & Jury of LLM & 50.00 & 38.71 & 24.80 \\
LEG Large & Jury of LLM & 52.08 & 40.57 & 26.70 \\
\midrule
LEG Base & Our Method & 60.33 & \textbf{60.69} & 36.12 \\
LEG Large & Our Method & \textbf{66.11} & 58.39 & \textbf{50.36} \\
\hline
\end{tabular}
\caption{Explanation classification performance on human-annotated test sets comparing jury-of-LLMs and our synthetic labeling method.}
\label{tab:jury-of-LLMs}
\end{table*}
\section{Human evaluation of synthetic explanation annotation quality}
\label{app:human-evaluation}
We conducted a human evaluation to assess the quality of the word-level annotations generated by \texttt{GPT-4o-mini}. One human expert was asked to label unsafe words in 50 randomly selected prompts from each test set: AEGIS2.0, WildGuardMix, and ToxicChat0124. We then compared these labels with the GPT-generated annotations using Cohen’s Kappa. The agreement scores were 54\% percent for AEGIS2.0, 54.50\% percent for WildGuardMix, and 43.56\% for ToxicChat0124, which indicate moderate agreement. We found that most disagreements were due to differences in phrase boundaries. For example, GPT often highlights shorter keywords like ``kill'' or ``harm'', while the human annotator marks longer phrases such as ``kill someone'' or ``cause harm to others''. In most cases, the core unsafe terms were present in both annotations.

\section{Generalization beyond synthetic supervision}
\label{app:generalization-of-explainability}

Since the explainability classifier of \name{} is trained using synthetic explanation labels generated by \texttt{GPT-4o-mini}, an important question is whether its explainability performance is upper-bounded by that of \texttt{GPT-4o-mini}, or whether \name{} can learn explainability behavior that generalizes beyond the synthetic supervision used during training. To examine this, we evaluated both \name{} and \texttt{GPT-4o-mini} against human-annotated gold labels, as described in Appendix~\ref{app:human-evaluation}. For \texttt{GPT-4o-mini}, we conducted a standard zero-shot prompting evaluation in which the model was asked to identify unsafe words or phrases in each input prompt. Table~\ref{tab:explainability_eval} reports the results of this evaluation.

\begin{table}[h]
\centering
\resizebox{\columnwidth}{!}{%
\begin{tabular}{lccc}
\hline
\textbf{Model} & \textbf{AEGIS 2.0} & \textbf{WildGuardMix} & \textbf{ToxicChat0124} \\
\hline
Zero-shot GPT-4o-mini & 53.28 & 54.87 & \textbf{50.96} \\
LEG base    & 60.33 & \textbf{60.69} & 36.12 \\
LEG large   & \textbf{66.11} & 58.39 & 50.36 \\
\hline
\end{tabular}%
}
\caption{Performance of explainability classification on the human-annotated test set, reported using F1 scores.}
\label{tab:explainability_eval}
\end{table}

On AEGIS2.0, \name{} base achieved an F1 score of 60.33\%, while \name{} large reached 66.11\%. On WildGuardMix, \name{} base obtained 60.69\%, and \name{} large achieved 58.39\%. On the more challenging ToxicChat0124 dataset, \name{} base achieved 36.12\%, and \name{} large reached 50.36\%. Overall, both \name{} base and \name{} large outperform \texttt{GPT-4o-mini} on AEGIS2.0 and WildGuardMix. On ToxicChat0124, \texttt{GPT-4o-mini} performs slightly better. However, \name{} large achieves comparable performance, with a difference of less than one percentage point. 
These results demonstrate that, although the explainability classifier in \name{} is trained using synthetic explanation labels generated by \texttt{GPT-4o-mini}, its performance is not upper-bounded by \texttt{GPT-4o-mini}. Instead, \name{} surpasses the performance of \texttt{GPT-4o-mini} in explainability classification in most cases. These results further confirm that \name{} learns explainability behavior that aligns closely with human judgment.

\begin{table*}[h]
\centering
\resizebox{\textwidth}{!}{
\begin{tabular}{lcc}
\hline
\textbf{Unsafe category} & \textbf{Prompt classification} & \textbf{Explainability classification} \\
\hline
Causing material harm by disseminating misinformation & 97.7 & 70.27 \\
Copyright violations & 87.3 & 47.7 \\
Cyberattack & 98.9 & 70.3 \\
Defamation encouraging unethical or unsafe actions & 98.9 & 69.9 \\
Disseminating false or misleading information, encouraging disinformation campaigns & 98.9 & 69.9 \\
Fraud assisting illegal activities & 89.9 & 76.8 \\
Mental health over reliance crisis & 94.4 & 70.4 \\
Others & 79.8 & 85.8 \\
Private information individual & 87.5 & 75.8 \\
Sensitive information organization government & 92.5 & 75.0 \\
Sexual content & 88.1 & 76.0 \\
Social stereotypes and unfair discrimination & 77.4 & 73.0 \\
Toxic language hate speech & 100.0 & 78.5 \\
Violence and physical harm & 98.7 & 67.6 \\
\hline
\end{tabular}
}
\caption{Performance of \name{} base on fine-grained safety categories of the WildGuardMix test set, reported using F1 score.}
\label{tab:fine_grained_safety}
\end{table*}
\begin{table*}[h]
\centering
\resizebox{\textwidth}{!}{
\begin{tabular}{lcccc}
\hline
\textbf{Unsafe category} 
& \multicolumn{2}{c}{\makecell{\textbf{Setting 1 performance}\\\textbf{(Seen topic)}}} 
& \multicolumn{2}{c}{\makecell{\textbf{Setting 2 performance} \\ \textbf{(Unseen topic)}}} \\
\cmidrule{2-5}
& Prompt & Explainability 
& Prompt & Explainability \\
\hline
Copyright violations 
& 87.3 & 47.7 & 62.2 & 40.4 \\
Disseminating false or misleading information, encouraging disinformation campaigns 
& 98.9 & 69.9 & 98.9 & 67.3 \\
Fraud assisting illegal activities 
& 89.9 & 76.8 & 88.9 & 74.1 \\
Violence and physical harm 
& 98.7 & 67.6 & 100.0 & 67.7 \\
\hline
\end{tabular}
}
\caption{Performance of \name{} on risk topics seen and unseen during training. Reported values are F1 scores for both prompt classification and explainability classification.}
\label{tab:category_exclusion}
\end{table*}
\section{Jury-of-LLMs vs. our method for synthetic explanation generation}
\label{app:jury-of-LLM}
One of the most common approaches for generating synthetic labels is to use a jury of LLMs. To compare our synthetic explanation labeling method (Section~\ref{sec:training-word-label-generation}) with this approach, we conducted an experiment using a jury of three LLMs. We used three open-source models, Llama-3.1-8B-Instruct, Mistral-7B-Instruct-v0.3, and Qwen3-8B, to generate explanation labels. We prompted the jury models to identify unsafe words. A word was labeled as unsafe only when all three models unanimously agreed. We then trained LEG using these jury-generated explanation labels and evaluated its performance on explanation classification using human-annotated test sets.

The results in Table~\ref{tab:jury-of-LLMs} show that explanation classification performance improves substantially when the model is trained on synthetic labels generated by our method, compared to labels produced by a jury of LLMs. This indicates that our confirmation-bias-based labeling approach aligns more closely with human judgments and achieves better generalization.

\section{Performance on fine-grained risk categories}
\label{app:fine-grained-performance}

The WildGuardMix dataset contains annotations for 14 fine-grained risk categories. Table~\ref{tab:fine_grained_safety} shows the performance of \name{} base across all fine-grained risk categories in the WildGuardMix test set. The results indicate that our method robustly learns a diverse set of safety risks, achieving consistently strong performance across categories. In particular, \name{} base achieves perfect performance on relatively easier categories such as \emph{toxic language and hate speech} (100\% F1) in prompt classification. At the same time, it maintains strong and reasonable performance on more challenging categories, including \emph{copyright violations} and \emph{social stereotypes and unfair discrimination}, for both prompt and explainability classification. These results highlight the ability of our method to handle both common and difficult safety risks in a unified framework.

\begin{table*}[t]
\centering
\resizebox{.7\textwidth}{!}{
\begin{threeparttable}
\begin{tabular}{l l c c c}
\hline
\textbf{Train Dataset} & \textbf{Model} & \textbf{Model Size} & \textbf{FPR} & \textbf{F1 score} \\
\hline
\multirow{5}{*}{--} 
 & Llama Guard 3$^{\dagger}$ & 1B &--& 43.4 \\
 & ShieldGamma$^{\dagger}$ & 2B &--& 69.4 \\
 & Llama Guard 2$^{\dagger}$ & 8B &--& 88.88 \\
 & Llama Guard 3$^{\dagger}$ & 8B &--& 88.4 \\
 & DuoGuard$^{\dagger}$ & 0.5B &--& 82.3 \\
\hline
\multirow{4}{*}{AEGIS2.0} 
& Prompt Baseline base & 86M &38.8 & 75.99\\
& Prompt Baseline large & 304M & 33.6 & 75.43\\
 & \name{} base & 86M & 40.0 & 75.0 \\
 & \name{} large & 304M & 22.0 & 77.99 \\
\hline
\multirow{4}{*}{WildGuardMix} 
& Prompt Baseline base & 86M & 5.6 & 87.07\\
& Prompt Baseline large & 304M & 6.4 & 88.08\\
 & \name{} base & 86M &5.6& 85.56 \\
 & \name{} large & 304M &\textbf{1.6}& \textbf{92.91} \\
\hline
\multirow{4}{*}{ToxicChat0124} 
& Prompt Baseline base & 86M &12.4& 67.24\\
& Prompt Baseline large & 304M &12.4& 69.12\\
 & \name{} base & 86M &6.8& 72.73 \\
 & \name{} large & 304M &11.2& 77.09 \\
\hline
\end{tabular}

\begin{tablenotes}[flushleft]
\footnotesize
\item[$\dagger$] F1 scores are reported as provided in \citep{deng2025duoguardtwoplayerrldrivenframework}.
\end{tablenotes}
\end{threeparttable}
}
\caption{Prompt classification false positive rate (FPR) and F1 score on XSTest across different training datasets and models.}
\label{tab:xstest_results}
\end{table*}

\section{Performance on unseen risk topics}
\label{app:unseen-category-performance}

To evaluate whether \name{} can generalize to unseen risk topics, we conduct an experiment using the WildGuardMix dataset. In addition to binary safety labels (safe vs.\ unsafe), WildGuardMix annotates unsafe prompts with 16 fine-grained risk topics (e.g., violence, copyright violations). For this experiment, we randomly select four risk topics from WildGuardMix and train a variant of \name{} after removing all training instances associated with these topics. We then compare its performance against a model trained on the full dataset.

Specifically, we consider the following two training settings:
\begin{enumerate}
\item \textit{Setting 1 (Full training):} The model is trained on the complete WildGuardMix training set, covering all risk topics.
\item \textit{Setting 2 (Topic-excluded training):} The model is trained on a subset of the training data that excludes all instances from four randomly selected risk topics (shown in Table~\ref{tab:category_exclusion}). As a result, prompts associated with these topics are entirely unseen during training.
\end{enumerate}

Importantly, in both settings, the model is trained and evaluated using the same binary prompt-level labels (safe vs.\ unsafe). The exclusion affects only the topical content of the training data, not the label space.

The objective of this experiment is to assess whether \name{} can correctly classify prompts from previously unseen risk topics as unsafe. Table~\ref{tab:category_exclusion} reports test-set performance for the four topics under both training settings. The results show that in three out of four cases, \name{} achieves performance comparable to the full-training setting, despite having no exposure to these topics during training. The only notable drop occurs for \textit{copyright violations}, where accuracy decreases to 62.2\%, which is still reasonable given that this topic is entirely unseen during training.

These results suggest that \name{} does not rely on memorizing topic-specific lexical cues. Instead, it learns higher-level semantic signals associated with unsafe behavior, enabling robust generalization to unseen risk categories.
\section{Evaluation on benign prompts containing harmful words}
\label{app:eval-benign-prompt}
XSTest \citep{rottger-etal-2024-xstest} is a stress-test benchmark designed to evaluate robustness against superficial lexical cues in safety classification. Unlike standard safety datasets, XSTest focuses on benign prompts that intentionally use words commonly associated with harmful content. For example, it includes prompts such as ``how to kill a Python process?'', which contain word like `kill' but express clearly non-harmful intent. As a result, strong performance on XSTest indicates that a model is less reliant on keyword-based heuristics and better able to reason about context and intent.

We evaluate the prompt classification performance of \name{}. Since this dataset does not include explainability annotations, we do not evaluate the explainability classifier on this benchmark. Table~\ref{tab:xstest_results} reports F1 scores on XSTest for several existing guardrail models and for \name{} trained on different datasets. Additionally, we report the false positive rate (FPR) for \name{} and the Prompt Baseline described in Section~\ref{sec:baselines}.

\name{} achieves strong performance on XSTest despite being orders of magnitude smaller than existing guardrail models. In particular, \name{} large trained on WildGuardMix achieve the highest F1 score (92.91), outperforming all evaluated baselines, including large instruction-tuned guardrails such as Llama Guard 2 (8B) and Llama Guard 3 (8B). Notably, this result is achieved with a model that is several orders of magnitude smaller in parameter count.

Even the base variant of \name{} (86M parameters) performs competitively, achieving an F1 score of 85.56 on XSTest. These results highlight that \name{} can effectively handle exaggerated safety scenarios and benign prompts containing harmful lexical cues. Further, \name{} achieves a lower FPR in most cases than the Prompt Baseline, indicating that joint training with both prompt-level and explanation-level supervision helps reduce false positives compared to a model trained with only prompt-level supervision. Overall, these results demonstrate that \name{} learns context-sensitive safety representations rather than relying on superficial lexical cues. Crucially, it achieves this robustness while remaining lightweight.
\end{document}